\documentclass[preprint,12pt]{elsarticle}

\usepackage{tikz}
\usepackage{comment}
\usepackage{amsmath,amssymb} % define this before the line numbering.
\usepackage{color}
\usepackage{graphicx}
\usepackage{lineno}

\begin{document}

\begin{frontmatter}

\title{Unlimited Knowledge Distillation for Action Recognition in the Dark}

\author[addr1]{Ruibing Jin}
\author[addr2]{Guosheng Lin}
\author[addr1]{Min Wu}
\author[addr1]{Jie Lin}
\author[addr1]{Zhengguo Li}
\author[addr1]{Xiaoli Li}
\author[addr1]{Zhenghua Chen\corref{CorrAuthor}}
\ead{chen0832@e.ntu.edu.sg}

\address[addr1]{A*STAR, Institute for Infocomm Research, Singapore 138632}
\address[addr2]{School of Computer Science and Engineering, Nanyang Technological University (NTU), Singapore 639798}

\begin{abstract}
Dark videos often lose essential information, which causes the knowledge learned by networks is not enough to accurately recognize actions. Existing knowledge assembling methods require massive GPU memory to distill the knowledge from multiple teacher models into a student model. In action recognition, this drawback becomes serious due to much computation required by video process. Constrained by limited computation source, these approaches are infeasible. To address this issue, we propose an \textbf{u}nlimited \textbf{k}nowledge \textbf{d}istillation (UKD) in this paper. Compared with existing knowledge assembling methods, our UKD can effectively assemble different knowledge without introducing high GPU memory consumption. Thus, the number of teaching models for distillation is \textbf{unlimited}. With our UKD, the network's learned knowledge can be remarkably enriched. Our experiments show that the single stream network distilled with our UKD even surpasses a two-stream network. Extensive experiments are conducted on the ARID dataset.
\end{abstract}


\end{frontmatter}

%\linenumbers

%% main text
\section{Introduction}

Action recognition has been widely studied in recent decades, where many approaches \cite{feichtenhofer2019slowfast,carreira2017quo,girdhar2017actionvlad,qiu2017learning,simonyan2014two,tran2018closer,wang2016temporal,wang2018non,yang2020temporal,li2020spatio,yang2019asymmetric,wang2013learning} have been proposed to investigate how to capture temporal features for the action classification. With the remarkable progress of action recognition, many relevant methods have been applied in the surveillance system. However, most existing approaches \cite{feichtenhofer2019slowfast,carreira2017quo,girdhar2017actionvlad,qiu2017learning,simonyan2014two,tran2018closer,wang2016temporal,wang2018non,yang2020temporal} are proposed for recognizing actions under normal environment. In comparison, crime often occurs under adverse conditions like night or occluded environment. Since video under adverse condition is significantly different from normal video, it is challenging to utilize existing methods to accurately classify actions under adverse condition. To solve this problem, we investigate how to effectively recognize actions under poor lighting conditions in this paper.

\begin{figure}[htbp]
	\begin{center}
		\includegraphics[width=.9\linewidth]{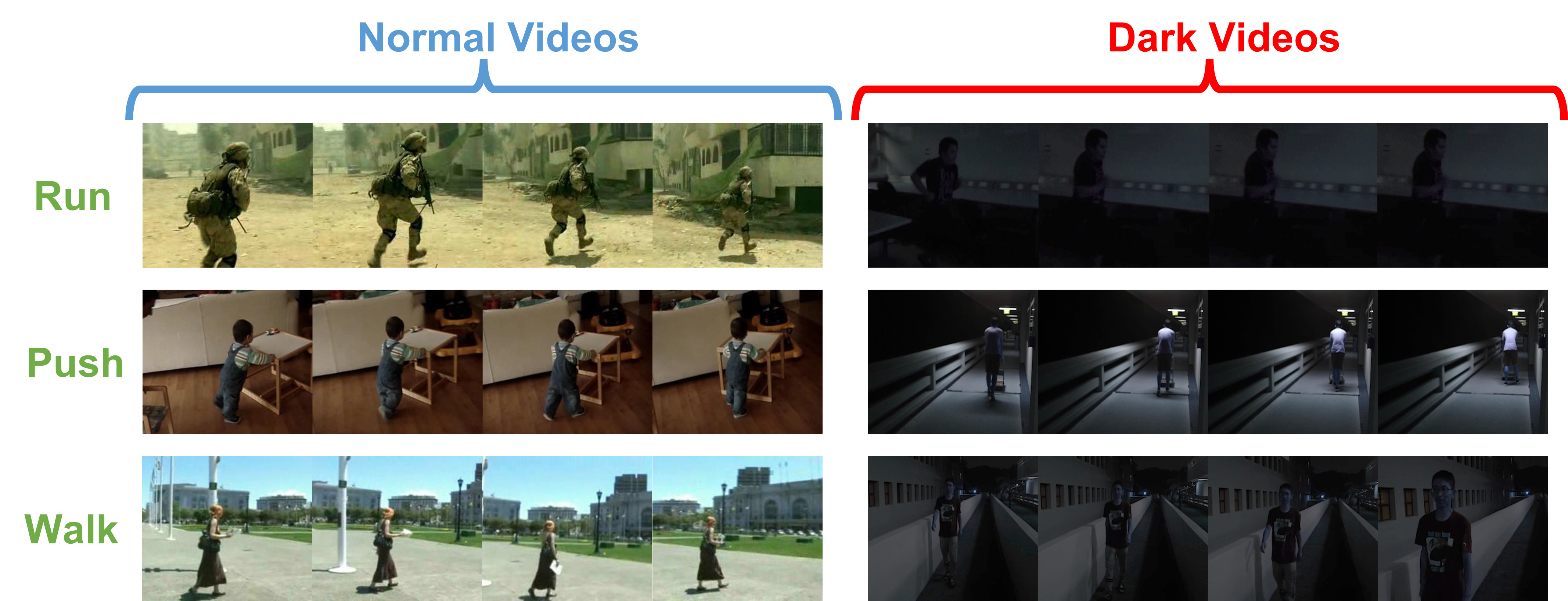}
	\end{center}
	\caption{Three normal and dark videos are illustrated. The normal videos are collect from HMDB51 dataset and the dark videos are collected from the ARID dataset. For a fair comparison, we divide them into three groups, when each group under the same action class. It can be found that dark videos are significantly different from normal videos. This increases the difficulty in action recognition.}
	\label{fig:vid_comp}
\end{figure}

Action recognition under dark is different from general action recognition in three aspects. As shown in Fig. \ref{fig:vid_comp}, six videos are illustrated, where three videos are captured under normal condition, while another three videos are recorded under dark. For a clear comparison, two videos on the same row belong to the same action category. It can be found that compared with normal videos, the quality of dark videos is seriously affected, leading to a significant loss of important visual information. Many important cues, which can be used to recognize actions in normal videos, cannot be captured by a neural network in dark videos. This makes it challenging to recognize actions under dark. Moreover, when video become dark, its color histogram changes dramatically \cite{xu2021arid}. This causes a domain gap between dark videos and normal videos. It is difficult to directly transfer learned knowledge from normal videos to these dark videos. Additionally, dark videos with tags are not widely available, and thus the number of these videos are limited. This further increases the difficulty in recognizing actions under dark. Constrained by these three factors, we investigate how to exploit more useful information for action recognition within a fixed number of dark videos in this paper.

\begin{figure}[htbp]
	\begin{center}
		\includegraphics[width=\linewidth]{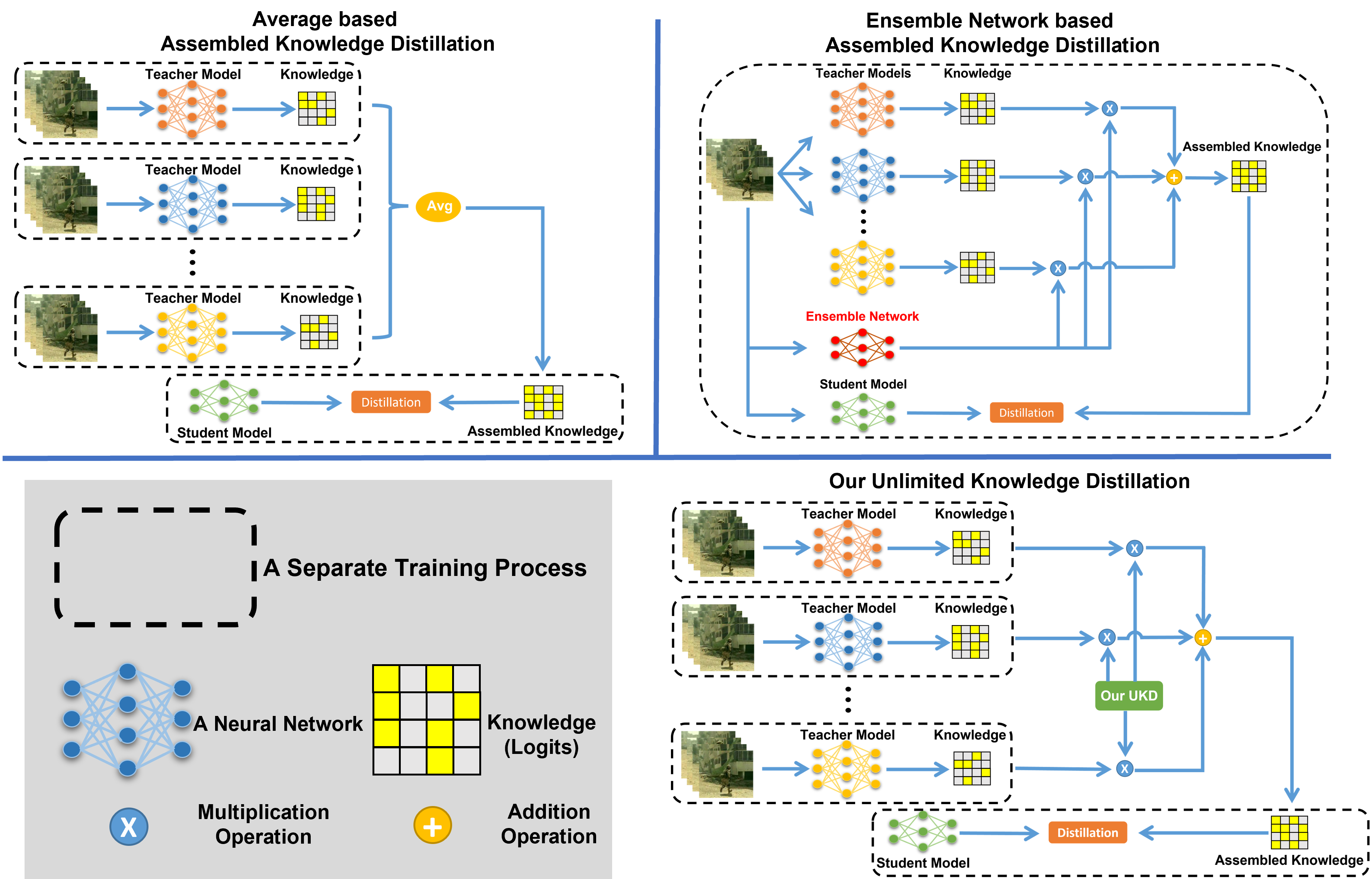}
	\end{center}
	\caption{Different assembled knowledge distillation methods. The average based assembled method, ensemble network based method and our unlimited knowledge distillation (UKD) method are illustrated at the top left, the top right and the lower part, respectively. Compared with other methods, our UKD can combine different knowledge without introducing additional networks in training. Our UKD is not limited by the computation source and the number of distilled knowledge is \textit{unlimited}.}
	\label{fig:kd_comp}
\end{figure}

In deep learning based methods, the exploited information in videos can be regarded as the knowledge learned by a neural network. So, we alternatively study how to enhance the learned knowledge by a neural network for improving the action recognition accuracy in dark videos. Knowledge distillation is widely used to transfer the teacher models' knowledge into a student model \cite{hinton2015distilling}. This distillation process can also be regarded as a kind of knowledge enhancement. To further enrich the knowledge, some methods \cite{tan2019multilingual,mirzadeh2020improved,tarvainen2017mean} are proposed to average multiple teachers' logits to improve the knowledge quality. Their process can be illustrated in the top left part of Fig. \ref{fig:kd_comp}. Different teacher models are firstly trained separately to produce the logits. Then, these logits are averaged to generate an assembled knowledge, which is distilled into a student model. Since each model training process is conducted separately, this average based assembled knowledge distillation does not cost massive GPU memory.

However, these approaches neglect the difference of importance among different logits, which limits the quality of the assembled knowledge. It is challenging to directly forecast the importance of different logits for a student model. To achieve this target, some methods \cite{xiang2020learning,zhang2018better,zhu2018knowledge} propose to utilize the neural network's adaptive learning property to online learn the knowledge weights. As shown in the top right of Fig. \ref{fig:kd_comp}, an ensemble network is proposed to dynamically weighted sum different teachers' knowledge together into a new knowledge. Although the difference of importance for different logits is captured, these approaches introduce additionally multiple teacher models in the training process, which causes that this kind of knowledge distillation occupies a large amount of GPU memory. In action recognition, 3D convolutional operation which is used for video processing, is computation expensive \cite{feichtenhofer2019slowfast,carreira2017quo}. Action recognition methods often suffer from the computation cost problem \cite{chen2018multi}. Thus, it is infeasible to adopt this ensemble network based knowledge distillation in action recognition. 

\begin{figure}[htbp]
	\begin{center}
		\includegraphics[width=.7\linewidth]{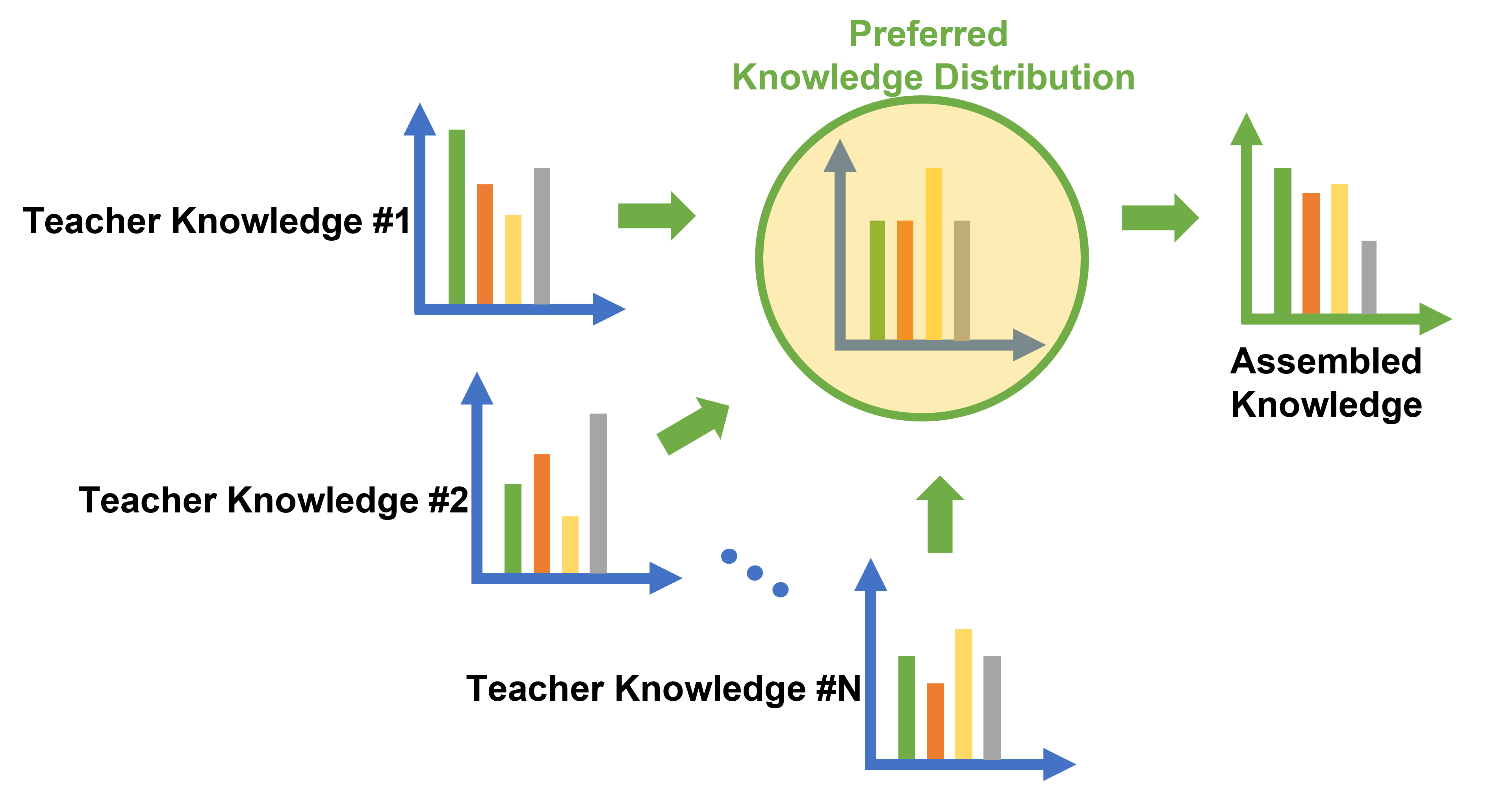}
	\end{center}
	\caption{The illustration of our preferred knowledge distribution (PKD). Different teacher knowledges are compared with our PKD. After that, the assembled knowledge is produced based on the comparison results.}
	\label{fig:pkd}
\end{figure}

To alleviate this issue, we propose a new knowledge distillation approach called \textbf{u}nlimited \textbf{k}nowledge \textbf{d}istillation (UKD), which is shown in the lower part of Fig. \ref{fig:kd_comp}. As illustrated in Fig. \ref{fig:kd_comp}, each teacher model is trained separately in our UKD. After the training process, our proposed UKD assembles these teacher knowledge offline. Then, we distill this assembled knowledge into a student model. Existing methods ignore how to evaluate the knowledge importance and simply utilize the neural network's adaptive learning property to predict the importance for different knowledge. Different from them, our UKD considers how to evaluate the importance for knowledge. In our UKD, we define a \textbf{p}referred \textbf{k}nowledge \textbf{d}istribution (PKD) based on \cite{yuan2020revisiting} which serves as a criterion, to evaluate the importance for different knowledge. The method \cite{yuan2020revisiting} uses this distribution for self-supervised knowledge distillation, while we develop it for knowledge ensemble. As shown in Fig. \ref{fig:pkd}, different teacher knowledges are compared with our PKD. Then, based on the comparison results, our UKD assigns different weights on different teacher knowledge. After that, our UKD produces the assembled knowledge via a weighted sum operation.

Compared with existing assembling knowledge distillation methods, our UKD is able to effectively assemble different knowledge together after the training process. It does not introduce any additional neural network into the training process and greatly alleviates the problem of computation cost. This characteristic enables our UKD to assemble different knowledge without the limitation of computation source. The number of assembled knowledge is \textit{unlimited}.

Recently, MARS \cite{crasto2019mars} is proposed to distill knowledge learned from a motion stream into a RGB frame based network. This method aims to enable the network based on RGB frames to mimic the motion stream and thus remove the computation cost on the motion stream. However, it only transfers the motion knowledge into a RGB frame based network and the knowledge ensemble is not involved. Different from MARS \cite{crasto2019mars}, our proposed UKD focuses on how to effectively assemble different knowledge.

To verify the effectiveness of our proposed UKD, extensive experiments are conducted on the ARID dataset, which is collected for action recognition under dark. It is used as a benchmark in the 4th $\rm{UG^{2}}$ Challenge in the CVPR 2021. Our experimental results demonstrate that our proposed UKD is able to effectively assemble different knowledge together. Distilled with this assembled knowledge, the action recognition accuracy in dark videos is improved. \textit{With our proposed UKD, a single stream network even surpasses a two-stream network for action recognition on dark videos.}

\section{Related Work}

\subsection{Action Recognition}

As an important task in video understanding, action recognition has been studied for several decades. Approaches in this task can be roughly divided into three categories: two-stream based method, 3D convolution based method and efficient computation method. Two-stream based methods generally consists of two neural networks, where one network are forwarded with RGB frames and another network extracts features from optical flow. Two-stream based methods focus on how to fuse these two neural networks for accurately classify actions. The two-stream network architecture is firstly proposed in \cite{simonyan2014two}. Following it, Fusion \cite{feichtenhofer2016convolutional} investigates different fusion modules to integrate the motion information with RGB frame information together. In TSN \cite{wang2015towards}, several practices are developed to improve the action recognition accuracy. ActionVLAD\cite{girdhar2017actionvlad} proposes a spatial and temporal aggregation scheme for some action primitives. Apart from it, some other approaches \cite{wang2015action,donahue2015long,kar2017adascan} are also proposed for better performances in action recognition.

Most two-stream based methods apply the 2D convolutional layer to capture the spatial and temporal information. For better performances, some approaches investigate how to extend the convolution operation from 2D to 3D. In 3D convolution based methods, C3D \cite{tran2015learning} shows that the 3D convolution layer is more suitable for action recognition than the 2D convolution layer. I3D \cite{carreira2017quo} rethinks the common architecture of action recognition methods and proposes a new 3D convolution based network. R3D \cite{hara2018can} successfully extends the 2D residual network architecture \cite{he2016deep} to the 3D space and applies this 3D residual network to action recognition. Non-local \cite{wang2018non} proposes a non-local operation for increasing the action recognition performance. SlowFast \cite{feichtenhofer2019slowfast} develops two pathways for processing videos with different frame rates. 

Although the 3D convolution layer can effectively capture the useful information in videos, its complex computation dramatically increases the computation cost. To solve this problem, several approaches like P3D\cite{qiu2017learning} and R2+1D \cite{tran2018closer} propose to replace the 3D operation with a 2D spatial convolution and a 1D temporal convolution. Following them, some methods \cite{wang2018appearance,chen2018multi,zhou2018mict} are proposed to optimize the network architecture for efficient computation. Although many methods are proposed for action recognition, most of them are proposed for videos under normal condition. Since dark video is different from normal video, their performances on dark video are not satisfactory. To solve this issue, we propose the unlimited knowledge distillation (UKD) for action recognition under dark.

\subsection{Knowledge Distillation}

Knowledge distillation currently has been widely used for transferring knowledge from a teacher model to a student model. It is firstly proposed in \cite{bucilua2006model} and then extended in \cite{hinton2015distilling}. In view of the distilled knowledge, the approaches for knowledge distillation can be roughly divided into two categories: logits based methods \cite{hinton2015distilling,tian2019contrastive,ba2014deep,cho2019efficacy,yang2019snapshot} and intermediate feature methods \cite{romero2014fitnets,heo2019comprehensive,kim2018paraphrasing,huang2017like}.

Conventional knowledge distillation methods only transfer the knowledge from a single teacher model into a student model. To further enrich the distilled knowledge, some approaches \cite{tan2019multilingual,mirzadeh2020improved,tarvainen2017mean,xiang2020learning,zhang2018better,zhu2018knowledge,wu2019distilled,park2019feed} are proposed to ensemble knowledge from multiple teacher models. Among these methods, some approaches\cite{tan2019multilingual,mirzadeh2020improved,tarvainen2017mean} are proposed to average the logits from multiple teacher models. Although these approaches are computation efficient, the difference of importance among different knowledge is not considered, which limits the quality of the assembled knowledge. Some methods \cite{xiang2020learning,zhang2018better,zhu2018knowledge} try to online compute weights for different knowledge. However, these methods often require to load additionally multiple neural networks during the training process. This causes that their computation cost is huge. Thus, it is challenging to apply these methods on the tasks which are computation expensive, like action recognition. Several approaches \cite{zhang2018better,zhu2018knowledge} are developed to assemble multiple feature based knowledge. Although their performances may be better, these approaches may occupy more GPU memory.

Existing knowledge distillation methods for knowledge ensemble cost much computation source. This drawback becomes serious when they are applied to the action recognition task, which often requires massive GPU memory for video processing. To solve this problem, we propose an unlimited knowledge distillation (UKD), which does not introduce any additional teacher model during the training process.

\section{Main Work}

In this section, we present our proposed unlimited knowledge distillation (UKD). In general, the distilled knowledge can be roughly divided into two categories: logits and intermediate feature representation. Since our aim is to reduce the computation cost for the assembled knowledge distillation, we choose to assemble logits or dark knowledge from different teacher models.

\subsection{Distillation from a single teacher model}

In video recognition, to distill the logits of a teacher model into a student model, the loss function can be formulated as:

\begin{equation}
	L = \alpha*L_{c}+ (1-\alpha)*L_{KL},
	\label{eq:total_loss}
\end{equation}where $L_{c}$ and $L_{KL}$ represent a standard cross entropy loss and a
knowledge distillation part, respectively. Given $N$ videos from $C$ classes, $L_{c}$ can be computed as following:

\begin{equation}
	L_{c} = -\frac{1}{N}\sum_{i=1}^{N}\sum_{j=i}^{C}y_{i,j}\log(p_{i,j}),
	\label{eq:basic}
\end{equation}where $y_{i,j}$ is the ground truth for class $j$ in video $i$, which is equal to 1 if video $i$ belongs to class $j$, and 0 otherwise. $p_{i,j} = {\rm softmax}(o_{i,j}) = \frac{exp(o_{i,j})}{\sum_{c=1}^{C}exp(o_{c,j})}$ indicates the prediction from a neural network for class $j$ in video $i$ and $o_{i,j}$ is the logits of a neural network. $L_{KL}$ is used to distill the knowledge of a teacher model into a student model. For a conventional knowledge distillation method, a Kullback Leibler (KL) Divergence is often used to compute the distance between two probability distributions, $p^{\tau}$ and $q^{\tau}$. This computation can be formulated as:

\begin{equation}
	L_{KL} = D_{KL}(q^{\tau}||p^{\tau})= \tau^{2}\frac{1}{N}\sum_{i=1}^{N}\sum_{j=1}^{C}q^{\tau}_{i,j}\log\frac{q^{\tau}_{i,j}}{p^{\tau}_{i,j}},
\end{equation}where $p^{\tau} = {\rm softmax}(o_{i,j}/\tau) = \frac{exp(o_{i,j}/\tau)}{\sum_{c=1}^{C}exp(o_{c,j}/\tau)}$ represents the probability of the soften logits from a student model and $q^{\tau}={\rm softmax}(z_{i,j}/\tau) = \frac{exp(z_{i,j}/\tau)}{\sum_{c=1}^{C}exp(z_{c,j}/\tau)}$ is the probability of the soften logits from a teacher model, where $z$ is the logits of the teacher model. $\tau$ indicates the temperature for label smooth.

\subsection{Average based Distillation from multiple teacher models}

To ensemble different knowledge, a common approach is to assign the same weight on multiple teacher's logits. Concretely, we can regard the knowledge distillation from multiple teacher models as a multi-task learning, which is defined in Eq. \ref{eq:avg_multi}. 

\begin{equation}
	L_{KL}= \frac{1}{K}\sum_{k=1}^{K}D^k_{KL}(q^{\tau}_{k}||p^{\tau})
	\label{eq:avg_multi}
\end{equation}

In Eq. \ref{eq:avg_multi}, the logits from $K$ teacher models are simultaneously transferred into a student model, where the importance for each teacher logit are the same. Another choice to distill multiple logits is to average the multiple teacher's logits and then distill this averaged logit into the student model. This process can be formulated as:  

\begin{equation}
	L_{KL}= D^k_{KL}(q^{\tau}_{avg}||p^{\tau}),
	\label{eq:avg_sing}
\end{equation}where $q^{\tau}_{avg} = \frac{1}{K}\sum_{k=1}^{K}q^{\tau}_{k}$ Although these two approaches can assemble multiple teachers' logits together, the difference of importance among different teachers' logits is not considered, which may affect the quality of the assembled knowledge. 

\subsection{Unlimited Knowledge Distillation}

\begin{figure}[htbp]
	\begin{center}
		\includegraphics[width=\linewidth]{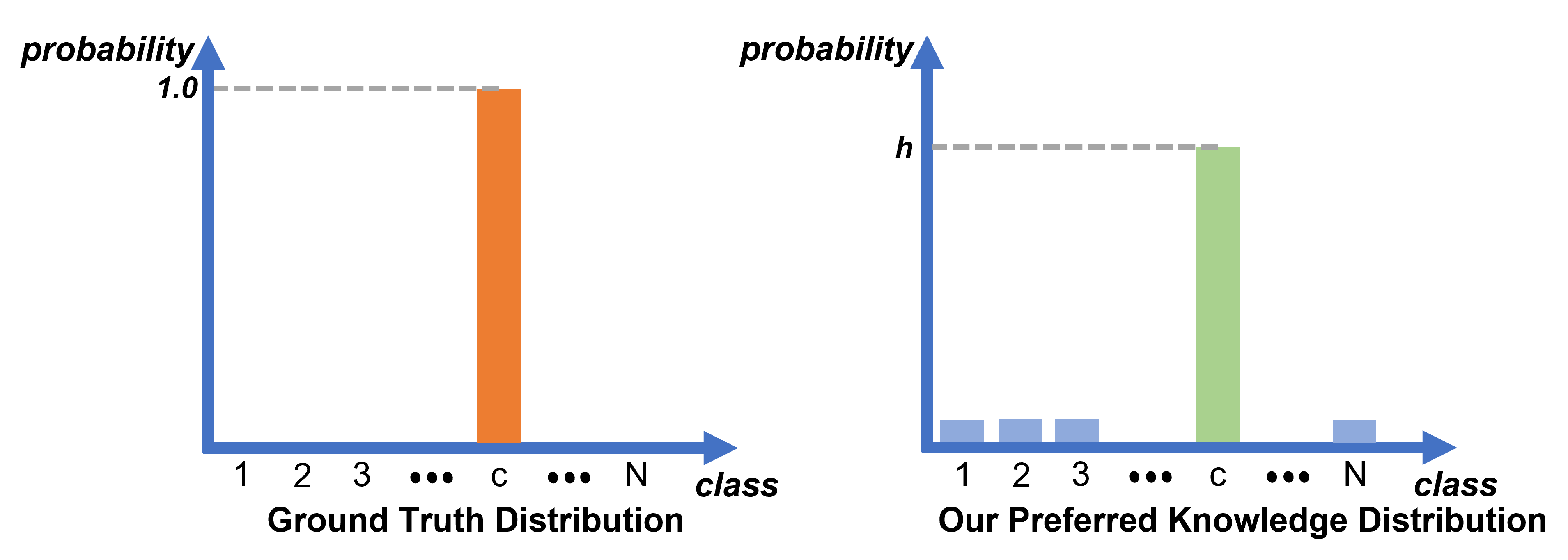}
	\end{center}
	\caption{Comparison between the ground truth distribution (GTD) and our preferred knowledge distribution (PKD). The GTD only sets the class $c$ as one, while neglects other classes. This kind of distribution suppresses the network learning on the relationship between classes, which may be conflict with the target of knowledge distillation. It is unreasonable to use GTD to evaluate the importance among knowledge. In comparison, our PKD encourages the network to learn the relationship between classes. This makes that our PKD is more suitable to evaluate the knowledge importance.}
	\label{fig:pkd2}
\end{figure}

To effectively distill the logits from multiple teachers, it is necessary to compute the importance for multiple teachers' logits. We hold an assumption that logits may include the informaiton on the relationship between classes. Based on our assumption, if a logit of a teacher model is correct, the class with the highest probability in this logit should be the ground truth. So, an intuitive method is to convert the ground truth into a ground truth distribution (GTD) and use this GTD to evaluate the importance of different logits, Then, the similarity $s_{n,k}$ between GTD and the logit of teacher model $k$ in video $n$, can be defined as follows:

\begin{equation}
	s_{n,k}= D^k_{KL}({\rm Y_{n}}||q^k_n)^{-1},
	\label{eq:w_1}
\end{equation}where $\rm{Y_{n}}$ indicates the probability distribution for the video $n$ ground truth, which is illustrated in the left side of Fig. \ref{fig:pkd2}. $q^k_{n}$ is the logits of teacher model $k$ in video $n$. 

To alleviate the computation burden, we can simplify the computation for $s_{n,k}$.  The $D_{KL}({\rm Y_{n}}||q^k_n)$ can be re-written as follows:

\begin{equation}
	\begin{aligned}
		D_{KL}({\rm Y_{n}}||q^k_n)& = \sum_{i=1}^{C}{\rm Y_{n}(i)}\log\frac{{\rm Y_{n}(i)}}{q^k_n(i)} \\
		& =\sum_{j=1}^{C}{\rm Y_{n}(i)}(\log({\rm Y_{n}(i)}) - \log(q^k_n(i))) \\
		& =\sum_{j=1}^{C}{\rm Y_{n}(i)}\log({\rm Y_{n}(i)}) - \sum_{j=1}^{C}{\rm Y_{n}(i)}\log(q^k_n(i))\\ 
	\end{aligned},
\end{equation}where ${\rm Y_{n}(i)}$ is a constant for our pre-defined probability distribution. Thus, the computation for $s_{n,k}$ can be written as:

\begin{equation}
	s_{n,k}= -\sum_{j=1}^{C}{\rm Y_{n}(i)}\log(q^k_n(i))^{-1} = {\rm CE}({\rm Y_{n}}, q^k_n)^{-1},
	\label{eq:w_2}
\end{equation}where ${\rm CE}({\rm Y_{n}}, q^k_n)$ represents the cross-entropy between the distribution $q^k_n$ relative to our pre-defined distribution ${\rm Y_{n}}$. After that, we normalize $\bar{s}_{n,k} = \frac{s_{n,k}}{\sum_{i=1}^{N}s_{n,i}}$. Then, the ensemble knowledge can be computed as:
\begin{equation}
	\bar{q}_n= \sum_{k=1}^{K}\bar{s}_{n,k}*q^k_n.
	\label{eq:p_1}
\end{equation}And the $L_{KL}$ is computed as:
\begin{equation}
	L_{KL}= D^k_{KL}(\bar{q}^{\tau}_{n}||p^{\tau}_{n}).
	\label{eq:l_1}
\end{equation}

However, this GTD based knowledge ensemble may not be reasonable. As shown in the left side of Fig. \ref{fig:pkd2}, in the GTD, only one class is assigned to one, while others are set as zero. This kind of probability distribution suppresses the network to learn the relationship between classes, which is conflict with the aim of knowledge distillation. To mitigate this issue, we propose a preferred knowledge distribution (PKD) based on \cite{yuan2020revisiting}. For a video in class $c$, our PKD is defined as following:
\begin{equation}
	q^{pkd}(x) = 
	\begin{cases}
		h &  x = c\\
		(1-h)/(C-1) & \rm{others} 
	\end{cases},
\end{equation}where $C$ is the class number and $h$ is a hyper-parameter.

As shown in the right side of Fig. \ref{fig:pkd2}, different from the GTD, our proposed PKD not only retains the ground truth information, but also encourages the neural network to learn the relationship between classes. This is consistent to the aim of the logit. With our proposed PKD, we can compute the weight for different teachers' logits according to Eq. \ref{eq:pkd_1}, and distill the assembled knowledge into a student model according to Eq. \ref{eq:p_1} and Eq. \ref{eq:l_1}.

\begin{equation}
	s_{n,k}= {\rm CE}(q^{pkd}, q^k_n)^{-1}
	\label{eq:pkd_1}
\end{equation}

\section{Experiments}

In this section, we introduce the dataset, the relevant implementation details and our experimental results.

\subsection{Experiment Settings}

\textbf{Dataset and evaluation metric.} In this paper, we use the ARID v1.5 to evaluate the performance of our proposed method. ARID \cite{xu2021arid} is collected for action recognition under dark environment and is used as a benchmark in the 4th $\rm{UG^{2}}$ Challenge in the CVPR 2021. Currently, there are three available versions, 1.0, 1.1 and 1.5 for this dataset. The ARID v1.5 adds more videos and more complex scenarios. There are 11 common classes such as walking, pushing, and turning in this dataset. The videos in this dataset are divided into two splits, where each split is further separated into a train set and a test set. In the split 0, there are 3350 videos in the train set and 2011 videos in the test set. In the split 1, there are 3792 videos in the train set and 1768 videos in the test set. For evaluation, we use the top-1 accuracy to evaluate the performances of methods.

\textbf{Implementation Details.} In our experiments, all the neural networks are pre-trained on the Kinetics-400 \cite{carreira2017quo}. The input frames are randomly cropped into 224 $\times$ 224 pixels with a shorter side randomly sampled in [224, 288]. The SGD is used to optimize our networks, where the learning rate is set as 0.01, batch size is 16 and the number of epoch is 30. For our PKD, the hyper-parameter $h$ is set as 0.99 and the $\alpha$ in Eq. \ref{eq:basic} is 0.5. The number of frame in training is 64 for the slowfast \cite{feichtenhofer2019slowfast} and 32 for other methods. 

Original dark videos lost much information for action recognition. To alleviate this issue, we apply the gamma intensity correction which is defined in Eq. \ref{eq:gamma}, according to ARID \cite{xu2021arid}. In Eq. \ref{eq:gamma}, $I(x, y)$ is the raw pixel value at $(x, y)$ in an image and $\bar{I}(x, y)$ represents the pre-processed value. All pixel values have been normalized in [0, 1]. The value for $\gamma$ is set as 3.0 in this paper.

\begin{equation}
	\bar{I}(x, y) = I(x, y)^{(1/\gamma)}
	\label{eq:gamma}
\end{equation}

\subsection{Ablation Study}

To verify the effectiveness of our proposed method, extensive experiments are conducted on the split 0 in ARID dataset. In this subsection, we re-implement the slowfast 4$\times$16 with Resnet50 according to \cite{feichtenhofer2019slowfast} and use it as our baseline.

Before knowledge ensemble, we firstly conduct experiments to investigate the effectiveness of the knowledge distillation with a single knowledge. The experimental results are listed in Table \ref{table:single}. RGB-logit indicates the logit of a RGB frame based teacher model, while flow-logit means the logit of an optical flow based teacher model. KD-RGB and KD-flow represent knowledge distillation with RGB-logit and flow-logit, respectively. Without knowledge distillation, we firstly train two single stream networks which are denoted as baseline-RGB and baseline-flow. As listed in Table \ref{table:single}, baseline-RGB performs much better than baseline-flow (67.4 v.s. 57.8), which demonstrates that RGB frames may be easier to capture useful information than optical flow for action recognition in dark videos.

\begin{table}[htbp]
	\caption{Comparison between methods distilled with a single knowledge on ARID aplit 0. The teacher model and the student model use the same network architecture. RGB-logit and flow-logit represent the distilled knowledge from the RGB based teacher model and the optical flow based teacher model, respectively.}
	\begin{center}
		\scalebox{.95}{
		\begin{tabular}{c|c|c|c|c}
			\hline
			Method &Input& Logit& $\tau$ & Top-1 Accuracy (\%)\\
			\hline
			baseline-RGB& RGB & N.A. & N.A. & 67.4 \\
			baseline-flow& flow & N.A. & N.A. & 57.8 \\
			\hline
			baseline-RGB+KD-RGB& RGB & RGB-logit & 5.0 & \textbf{70.8} \\
			baseline-RGB+KD-flow& RGB & flow-logit & 30.0 & 69.5 \\
			\hline
			baseline-flow+KD-flow& flow & flow-logit & 20.0 & 62.7 \\
			baseline-flow+KD-RGB& flow & RGB-logit & 10.0 & 62.1 \\
			\hline				
		\end{tabular}}
	\end{center}
	\label{table:single}
\end{table}

After that, to investigate the effectiveness of different logits, we use the logits from baseline-RGB and baseline-flow for knowledge distillation. We try to separately distill both different logits into the baseline-RGB model, and the trained models are represented by baseline-RGB+KD-RGB and baseline-RGB+KD-flow. Although the performance of baseline-RGB is greatly better than that of baseline-flow, the performance gap between baseline-RGB+KD-RGB and baseline-RGB+KD-flow is not obvious. This indicates that the teacher model performance may not be an effective criterion to evaluate the value for the logit. Since the RGB-logit and flow-logit are produced from different input modality, we believe that knowledge from different modalities may be complimentary. Then, we also distill these two logits into the baseline-flow, which are denoted as  baseline-flow+KD-flow and baseline-flow+KD-RGB. As listed in Table \ref{table:single}, the baseline-flow is significantly improved after knowledge distillation with RGB-logit and flow-logit.  

According to experiments above, the action recognition can be effectively improved by distilling the logit of a single teacher model. For better performances, we try to use the assembled knowledge for knowledge distillation. In Table \ref{table:abl}, we conduct an ablation study for four different knowledge ensemble methods. AVG-1 indicates the averaged based knowledge distillation method in Eq. \ref{eq:avg_multi}, and AVG-2 is the method defined in Eq. \ref{eq:avg_sing}. GTD means that we use our proposed GTD to compute the weights for different knowledge (flow-logit and RGB logit) ensemble and UKD represents that we use our proposed PKD to assemble flow-logit and RGB logit. 

\begin{table}[htbp]
	\caption{Ablation study for different knowledge ensemble methods on ARID split 0. AVG-1 indicates the averaged based method defined in Eq. \ref{eq:avg_multi}. AVG-2 is the method based on Eq. \ref{eq:avg_sing}. This experiment shows that our proposed UKD effectively assembles different logits and achieves the best performance.}
	\begin{center}
		\scalebox{0.85}{
		\begin{tabular}{c|c|c|c|c}
			\hline
			Method &Input& Logit& $\tau$ & Top-1 Accuracy (\%)\\
			\hline
			baseline-RGB& RGB & N.A. & N.A. & 67.4 \\
			baseline-flow& flow & N.A. & N.A. & 57.8 \\
			baseline-RGB+KD-RGB& RGB & RGB-logit & 5.0 & 70.8 \\
			\hline
			baseline+AVG-1& RGB & flow-logit+RGB-logit & 10.0 & 70.8 \\
			baseline+AVG-2& RGB & flow-logit+RGB-logit & 60.0 & 70.1 \\
			\hline
			baseline+GTD& RGB & flow-logit+RGB-logit & 20.0 & 68.7 \\
			baseline+UKD (ours) & RGB & flow-logit+RGB-logit & 20.0 & \textbf{71.3} \\
			\hline					
		\end{tabular}}
	\end{center}
	\label{table:abl}
\end{table}

In Table \ref{table:abl}, it can be found that two average based knowledge ensemble methods show similar performances to baseline-RGB+KD-RGB. This demonstrates that these two knowledge ensemble methods do not effectively assemble knowledge, since these two methods neglect the difference of importance among different knowledge. Different from these two methods, baseline+GTD and baseline+UKD propose to assign different weights on multiple knowledges. As shown in Table \ref{table:abl}, baseline+GTD performs inferior to baseline-RGB+KD-RGB and performs better than baseline-RGB. This shows that GTD may degenerate the quality of the logit, which verifies our hypothesis that GTD may suppress the network learning on the class relationship and is conflict with the aim of knowledge distillation. In comparison, our UKD utilizes our PKD to assign different weights on multiple logits and shows the best performance, 71.3. This demonstrates that our proposed UKD is able to effectively ensemble multiple teachers' logits.

\subsection{Comparison with other methods} 

For a comprehensive comparison with other methods, we re-implement some classic approaches in action recognition. According to Non-local \cite{wang2018non}, we re-implement I3D with non-local blocks based on ResNet-50 and Resnet-101, which are denoted as NL-I3DRes50 and NL-I3DRes101, respectively. We also re-implement the CSN \cite{tran2019video} with ResNet-152 and denotes it as IR-CSNRes152. TPN \cite{yang2020temporal} based on ResNet-50 is also re-implemented by us.  

\begin{table}[htbp]
	\caption{Comparison experiments with other methods on two ARID splits. The Top-1 Accuracy is the average on two splits. It shows that our single-stream based UKD achieves the best performances in these approaches and even surpasses the two-stream based slowfast method.}
	\begin{center}
		\begin{tabular}{c|c}
			\hline
			Method & Top-1 Accuracy (\%)\\
			\hline
			slowfast(baseline)-RGB & 68.6 \\
			slowfast(baseline)-flow & 60.0\\
			NL-I3DRes50-RGB & 51.9\\
			NL-I3DRes50-flow & 52.5\\
			NL-I3DRes101-RGB & 55.9\\
			NL-I3DRes101-flow & 53.4\\
			IR-CSNRes152-RGB & 64.3\\
			IR-CSNRes152-flow & 60.8\\
			TPNRes50-rgb & 60.2\\
			TPNRes50-flow & 57.7\\
			\hline
			\hline
			slowfast-twostream & 70.9 \\
			UKD (ours) & \textbf{71.3} \\
			\hline
		\end{tabular}
	\end{center}
	\label{table:comp}
\end{table}

In Table \ref{table:comp}, we conduct experiments on both two ARID splits and report the averaged top-1 accuracy. It can be found that our proposed UKD shows the best performances (71.3) among all methods. We also conduct an experiment for a two-stream network based on slowfast, which averages the predictions from two streams. It is surprising that our UKD which consists of a single stream network, even performs better than this two-stream based method. This indicates that our proposed UKD is able to achieve better performance with less computation cost.

\subsection{Computation Cost Analysis} 

In this subsection, we analyze the additional computation cost introduced by our proposed UKD during the training process. All experiments are conducted in the same condition. The experimental results are listed in Table \ref{table:cost}, where the slowfast is slowfast 4$\times$16 trained with 16 batch size and 64 RGB frame input. 

\begin{table}[htbp]
	\caption{Comparison for Computation Cost. Compared with single knowledge distillation, our proposed UKD does not increases any computation cost.}
	\begin{center}
		\scalebox{.85}{
		\begin{tabular}{c|c|c}
			\hline
			Method & GPU Memory & Time per epoch (second)\\
			\hline
			slowfast(baseline)-RGB & 13,697 MB & 312\\
			slowfast(baseline)-RGB + KD-RGB & 14,965 MB & 385 \\
			slowfast(baseline)-RGB + UKD (ours) & 14,965 MB & 385\\
			\hline
		\end{tabular}}
	\end{center}
	\label{table:cost}
\end{table}

We conduct an experiment denoted as slowfast(baseline)-RGB + KD-RGB, where we only distill knowledge from the logit of a single teacher to the student model. According to Eq. \ref{eq:total_loss}, since the $L_{KL}$ is introduced, the computation cost for slowfast(baseline)-RGB + KD-RGB is increased slightly. Then, we conduct the experiment, slowfast(baseline)-RGB + UKD, which uses our UKD to assemble RGB-logit and flow-logit for knowledge distillation. In Table \ref{table:cost}, it can be seen that our proposed UKD does not introduce any additional computation cost compared with the single knowledge distillation slowfast(baseline)-RGB + KD-RGB. This demonstrates that our proposed UKD can effectively assemble multiple logits, while does not increases much computation cost.

\section{Conclusion}

In this paper, we have proposed an unlimited knowledge distillation (UKD) method for action recognition under dark. In our UKD, we develop a preferred knowledge distribution (PKD) to effectively assemble multiple logits without increasing much computation cost. Extensive experiments have been carried out, which shows that compared with existing assembling knowledge distillation methods, our proposed UKD is able to effectively increase the action recognition accuracy under dark, while does not introduce much computation cost.

\bibliographystyle{elsarticle-num} 
\bibliography{egbib}

\end{document}